\RequirePackage[svgnames]{xcolor}

\documentclass[11pt,a4paper]{mystyle}

\usepackage[all]{hypcap}
\usepackage[svgnames]{xcolor}
\usepackage[comma,authoryear,compress]{natbib}
\usepackage{amsmath,amsfonts}
\usepackage{algorithmic}
\usepackage{algorithm}
\usepackage{array}
\usepackage{textcomp}
\usepackage{stfloats}
\usepackage{url}
\usepackage{verbatim}
\usepackage{graphicx}
\usepackage[font=small,labelfont=bf]{caption}
\usepackage{subcaption}
\usepackage{textcomp}
\usepackage{stfloats}
\usepackage{amssymb}
\usepackage{latexsym}
\usepackage[utf8]{inputenc}
\usepackage{microtype}
\usepackage{inconsolata}
\usepackage{ragged2e}
\usepackage{balance}
\usepackage{multirow}
\usepackage{multicol}
\usepackage{makecell}
\usepackage{tabularx,booktabs}
\usepackage{enumitem}
\usepackage{xspace}
\usepackage{fontawesome5}
\usepackage{float}
\usepackage{pifont}
\usepackage{microtype}
\usepackage{ulem}
\usepackage{bm}
\usepackage{autonum}
\usepackage{colortbl}
\usepackage{longtable}
\usepackage{hyperref}
\usepackage{wrapfig}
\usepackage{tcolorbox}
\usepackage{floatflt}
\usepackage{arydshln}
\usepackage{lineno}

\usepackage{dsfont}
\usepackage{cleveref}
\usepackage{bxcoloremoji}

\usepackage{xspace}
\usepackage{tcolorbox}
\usepackage{multicol}
\usepackage{multirow}
\usepackage{tabularx}
\usepackage{adjustbox}
\usepackage{bbding}
\usepackage{enumitem}
\usepackage{balance}
\usepackage{pifont}
\usepackage{graphicx}
\usepackage{array}
\usepackage{color}
\usepackage{booktabs}
\usepackage{arydshln}
\usepackage{rotating}

\definecolor{lightgrayv}{HTML}{F4F3F8} 
\definecolor{lightbluev}{HTML}{F5F9FD} 
\definecolor{grayv}{HTML}{707070}
\definecolor{bettergreen}{rgb}{0.13, 0.55, 0.13}
\definecolor{bluev}{HTML}{0070C0}
\definecolor{myblue}{HTML}{3889fe}
\definecolor{myorange}{HTML}{b2561a}
\definecolor{mygreen}{HTML}{6ea13f}

\newcommand{\modelname}{\textsc{InSide}\xspace}

\title{Bridging Thoughts and Words:\\Graph-Based Intent-Semantic Joint Learning for Fake News Detection}

\runningtitle{Bridging Thoughts and Words: Graph-Based Intent-Semantic Joint Learning for Fake News Detection}

\author[1,2]{\href{https://scholar.google.com/citations?user=fMCBAt4AAAAJ}{\textcolor{black}{Zhengjia Wang}}}
\author[1]{\href{https://sheng-qiang.github.io/}{\textcolor{black}{Qiang Sheng}}}
\author[1]{\href{https://scholar.google.com/citations?user=hGZwK0cAAAAJ}{\textcolor{black}{Danding Wang}}}
\author[1,2]{\href{https://scholar.google.com/citations?user=vPCpi_kAAAAJ}{\textcolor{black}{Beizhe Hu}}}
\author[1,2]{\href{https://scholar.google.com/citations?user=fSBdNg0AAAAJ}{\textcolor{black}{Juan Cao}}}

\affil[1]{Media Synthesis and Forensics Lab, Institute of Computing Technology, Chinese Academy of Sciences}
\affil[2]{University of Chinese Academy of Sciences}

\correspondingauthor{Juan Cao}
\emails{
    {\{\href{mailto:wangzhengjia21b@ict.ac.cn}{wangzhengjia21b},\href{mailto:shengqiang18z@ict.ac.cn}{shengqiang18z},\href{mailto:wangdanding@ict.ac.cn}{wangdanding},\href{mailto:hubeizhe21s@ict.ac.cn}{hubeizhe21s},\href{mailto:caojuan@ict.ac.cn}{caojuan}\}@ict.ac.cn}
}

\begin{document}
\begin{abstract}
Fake news detection is an important and challenging task for defending online information integrity.
Existing state-of-the-art approaches typically extract news semantic clues, such as writing patterns that include emotional words, stylistic features, etc.
However, detectors tuned solely to such semantic clues can easily fall into surface detection patterns, which can shift rapidly in dynamic environments, leading to limited performance in the evolving news landscape.
To address this issue, this paper investigates a novel perspective by incorporating news intent into fake news detection, bridging intents and semantics together.
The core insight is that by considering news intents, one can deeply understand the inherent thoughts behind news deception, rather than the surface patterns within words alone.
To achieve this goal, we propose Graph-based \textbf{In}tent-\textbf{S}emantic Jo\textbf{i}nt Mo\textbf{de}ling (\modelname) for fake news detection, which models deception clues from both semantic and intent signals via graph-based joint learning.
Specifically, \modelname reformulates news semantic and intent signals into heterogeneous graph structures, enabling long-range context interaction through entity guidance and capturing both holistic and implementation-level intent via coarse-to-fine intent modeling.
To achieve better alignment between semantics and intents, we further develop a dynamic pathway-based graph alignment strategy for effective message passing and aggregation across these signals by establishing a common space.
Extensive experiments on four benchmark datasets demonstrate the superiority of the proposed \modelname compared to state-of-the-art methods.
\vspace{15pt}

\coloremojicode{1F4C5} \textbf{Date}: September 2, 2025

\coloremojicode{1F4AC} \textbf{Venue}: The 34th ACM International Conference on Information and Knowledge Management\\Proceedings (CIKM'25)
\end{abstract}

\maketitle


\begin{figure}[htbp]
\centering
    \includegraphics[width=.65\linewidth]{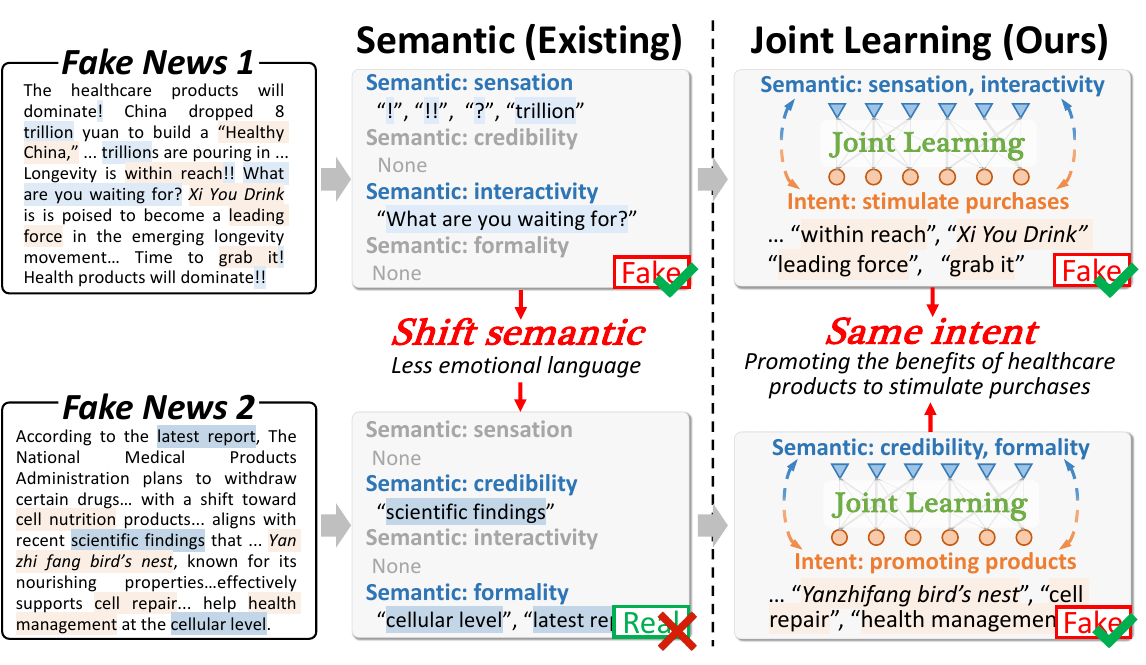}
    \caption{Comparison between existing works and our proposed framework. Existing detectors focus on surface semantic clues of news and often struggle when the writing pattern shifts. Instead, our method utilizes semantic-intent joint learning, integrating the thoughts (intents) and words (semantics) for enhanced detection. \label{fig:motive}}
\end{figure}

\section{Introduction}
The advancement of social media fundamentally transforms the way people access and consume information, facilitating the rapid dissemination of news worldwide \citep{sun2023fighting}.
However, such digital transformation also leads to the widespread proliferation of fake news, which exacerbates social polarization, manipulates public opinion, and undermines trust in reputable sources \citep{zhou2020survey}.
The widespread impact of fake news inflicts significant harm to society, including election integrity \citep{grinberg2019fake}, economic stability \citep{bollen2011twitter, arcuri2023does}, and democracy \citep{gillani2018me, bakir2018fake}.
Therefore, the development of effective automated approaches for detecting fake news has attracted significant attention \citep{ma2024fake, hu2024bad}, aiming to mitigate its harmful impact as early as possible.

Considering the fact that fake news is intentionally designed with specific patterns (\textit{e.g.,} emotionally charged language or specific narrative features) to attract and deceive the public, most existing detection methods typically focus on utilizing such semantic clues to detect fake news.
For example, \citet{sheng2021integrating} capture common stylistic patterns (\textit{e.g.,} emotional and negation words) to aid in fake news detection. Similarly, \citet{lutz2024linguistic} analyze the linguistic characteristics of fake news (\textit{e.g.,} adverbs, personal pronouns) and investigate how these cues contribute to its persuasiveness.
While such semantic clues play a vital role in identifying fake news, relying solely on them may lead detectors to overfit to static, surface-level writing patterns \citep{wu2024fake}. This limitation is particularly pronounced in today’s news landscape, where writing styles are diverse and constantly evolving—both over time \citep{zhu2022generalizing, hu2023learn, sheng2022zoom} and across domains \citep{nan2022improving, arora2023detecting}—posing new challenges for fake news detection systems.

To address this issue, cutting-edge approaches recommend combining external stable information beyond news content to assist the detection process, such as using user comments \citep{nan2024let} or propagation structure \citep{yin2024gamc, cui2024propagation, zhu2024propagation} as societal judgment, or exploiting factual evidence as external verification \citep{hu2021compare, zhang2024multi}.
While integrating such information can enhance detection performance, these methods face practical limitations in real-world scenarios, where societal feedback or relevant factual evidence may be unavailable or can only be gathered after the fake news has already spread widely and caused negative societal impact.
\textbf{This motivates us to investigate whether effective detection can be achieved from a stable and inherent perspective of news content itself.}

To this end, we observe that news intent offers an alternative perspective for achieving this objective.
News intent, defined as the purpose or motivation behind a news article, represents the fundamental characteristics of news articles \citep{zhou2020survey, bratman1984two, wang2025exploring}.
The basic idea is that the variability of fake news writing patterns stems from the inherent purpose of realizing the underlying thoughts, \textit{i.e.}, harmful intent, which in turn provides a stable perspective behind diverse choices of words in news.
For instance, as shown in \figurename~\ref{fig:motive}, the first news item employs writing patterns characterized by sensation (repeated words, \textit{e.g.,} ``!'' and ``trillions'') and interactivity (\textit{e.g.,} ``what are you waiting for?''), while the second one favors credibility (such as claims of ``scientific findings'' and ``cellular level'') and formality (phrases like ``latest health data'').
Despite their differences in semantic clues, they share the same intent, {\em i.e.,} ``promoting the benefits of healthcare products to stimulate purchases'', which can serve as a strong and stable signal for fake news detection.

Therefore, leveraging both semantic and intent signals provides a promising direction for deriving more expressive and deception-aware news representations.
However, how to use these signals effectively is nontrivial due to several key challenges:
\textbf{(i) For semantic modeling}, news contains multi-level relationships that holistic modeling approaches cannot adequately capture. Effectively modeling both local sentence-level interactions and global entity-referencing connections is crucial for understanding how intent is implemented through news narratives.
\textbf{(ii) For intent modeling}, straightforward classification-based solutions constrain representational capacity. An expanded intent representation space is needed to capture diverse intent manifestations across varied news contexts.
\textbf{(iii) For cross-signal integration}, semantic and intent signals have inherent representational differences; an effective alignment strategy is essential to bridge this gap and fully leverage their combined potential for fake news detection.

Based on the above discussions, we present a graph-based \textbf{in}tent-\textbf{s}emantic jo\textbf{i}nt mo\textbf{de}ling method, named \textbf{\modelname}, which bridges thoughts and words for fake news detection.
Specifically, for semantic modeling, we split the news into sentences and introduce involved entities to formulate a semantic heterogeneous graph.
The benefit is that we can explicitly model the specific dependencies via concatenated edges between sentence and entity nodes.
Accordingly, \modelname applies sliding window-based interaction for local sentence structures while leveraging entity connections for long-distance context interaction.
For intent modeling, \modelname utilizes large language models (LLMs) to extract unstructured news intents, expanding the intent representation space beyond fixed classifications. 
Additionally, considering the importance of intent implementation, \textit{i.e.}, how intent is expressed through news narrative in the context of fake news detection, we develop a coarse-to-fine news intent modeling strategy to enhance the correlation between intent and news narrative characteristics.
Besides, to extract representations from semantic and intent signals, \modelname performs dual-level graph updating for both semantic and intent graphs, enabling information exchange at local and global levels.
For cross-signal integration, \modelname proposes a dynamic pathway-based alignment module with pseudo nodes to bridge the semantic-intent gap through bidirectional message passing and aggregation, ultimately yielding effective representations for fake news detection.

Extensive experiments across four datasets demonstrate that our proposed \modelname consistently outperforms state-of-the-art methods. Beyond performance improvements, our work provides new insights into the crucial role of intent modeling in fake news detection, opening promising directions for future research.

\section{Related Work}

\subsection{Intent-agnostic Fake News Detection}
Intent-agnostic fake news detection methods focus on linguistic patterns, stylistic features, or social context without modeling the underlying intent of news articles.
More recent methods address this limitation by incorporating external information, such as social context (\textit{e.g.,} user profiles \citep{shu2019role, sitaula2020credibility}, user comments \cite{Shu_WeakSupervision, Ma_RumorStance}, and propagation structures \cite{sun2023fighting}) or factual evidence \cite{hu2021compare, dun2021kan, xu2022evidence, xie2023heterogeneous}. Although these approaches have shown promising results, their reliance on auxiliary sources remains a key limitation. When such information is scarce, particularly in the early stages of fake news dissemination, these methods may not fully exploit their advantages. 
To address this, we enhance content-based methods by extracting richer signals directly from news content, which can also complement external approaches to further improve detection performance.
Structure-aware methods explicitly model relationships within or between news articles. Some approaches construct graphs at different granularities: word-level methods \citep{yao2019graph, li2023grenade, gong2024heterogeneous, wang2024style} model keyword relationships to capture semantic dependencies, while syntax-based methods \citep{feng2012syntactic, perez2015experiments} leverage parsing tools to construct hierarchical textual structures.

Many existing structure-aware methods \citep{cui2024propagation, zhu2024propagation, yin2024gamc, FANGGraph} build graphs across multiple news articles or incorporate social context. However, these methods face two key limitations: (i) they rely on sufficient social context, making them less effective for emerging news events; (ii) they often introduce inter-article connections that are unavailable in real-time detection, resulting in a semi-supervised setting that may not reflect practical scenarios.

In contrast to approaches dependent on social context or external knowledge, \modelname builds graphs from individual news articles rather than across multiple ones, making it well-suited for early detection of newly emerging news.

\begin{figure*}[ht]
    \centering
    \includegraphics[width=1.\linewidth]{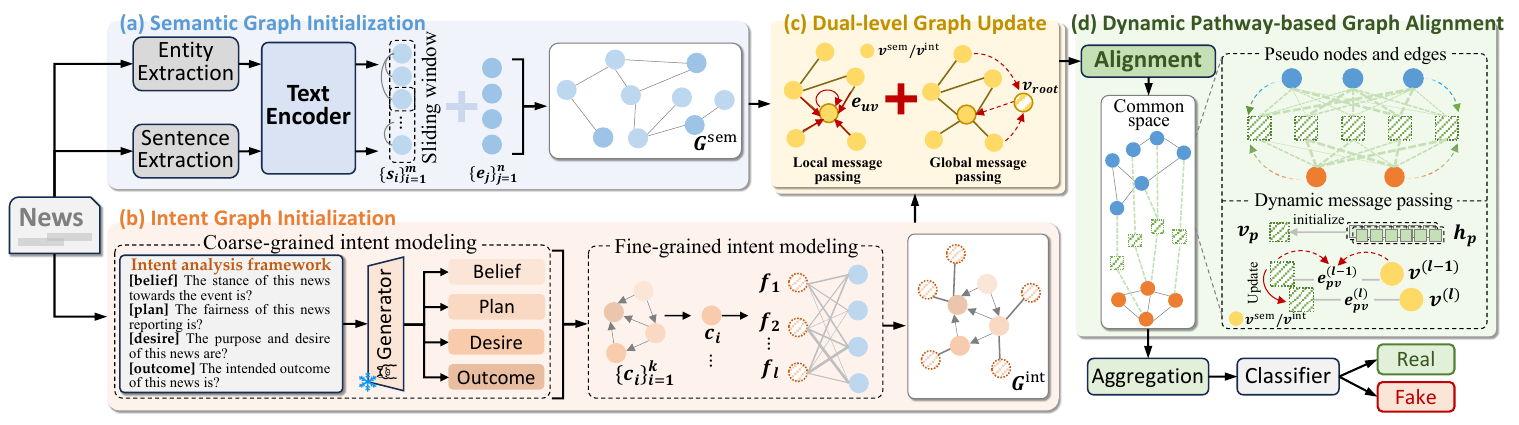}
    \caption{\label{fig:method} Overview of \modelname architecture. Given a news item, \modelname leverages heterogeneous graph structures to represent the news semantics and intent (a-b). 
    After obtaining the encoded semantic and intent representations via dual-level graph updating (c), a dynamic pathway-based graph alignment strategy is utilized to enable cross-signal alignment within a common space (d). The illustrative intent analysis framework is provided as an example and can be replaced with alternative ones.}
\end{figure*}

\subsection{Intent-aware Fake News Detection}
Intent-aware methods recognize the importance of modeling news creation intents behind news content, which provides crucial signals for distinguishing deceptive content.
Early approaches implicitly model news intent through social engagement patterns. For instance, some methods \citep{shu2019role, sitaula2020credibility, Shu_WeakSupervision, Ma_RumorStance} analyze user comments or news propagation patterns to implicitly model news creators' intents through mass reactions. While these methods provide valuable insights, they rely on auxiliary information that might be unavailable during the early stages of news propagation, limiting their practical applicability for timely detection.

The explicit modeling of news intent has recently emerged as a promising direction. \citet{wang2025exploring} draw from interdisciplinary intent theory research and apply it to the news domain, establishing a systematic framework for news intent analysis. While not specifically designed for fake news detection, this work demonstrates the potential of intent modeling for distinguishing deceptive content. 
\citet{wang2024misinformation} incorporate intent signals using a classification-based approach. However, this method uses fixed categorical representations for intent modeling and employs unidirectional fusion from intent to semantic features, limiting its ability to capture nuanced intent interactions with semantic features.

Our proposed method \modelname advances intent-aware fake news detection through three key innovations: 
(i) Unlike methods treating semantic and intent as holistic representations, we formulate both signals as heterogeneous graph structures to capture multi-level internal relationships;
(ii) Rather than fixed classification-based intent modeling, we employ a coarse-to-fine strategy that leverages unstructured intent extraction to expand representation capacity;
(iii) We develop a dynamic pathway-based alignment module enabling bidirectional information exchange, mitigating the overlooked representation gap between semantic and intent signals, to fully leverage their combined potential for fake news detection.
As demonstrated in Section~\ref{sec:exp4.2}, these innovations allow \modelname to outperform existing methods across diverse benchmarks.

\section{Proposed Method: \modelname}

\subsection{Preliminaries} \label{sec:preliminary}
In this paper, we formulate fake news detection as a binary classification task based on semantic and intent signals.  
Given a news article, \modelname transforms it into two heterogeneous graphs that capture its semantic and intent representations, denoted as $\mathbf{G}^\text{sem}$ and $\mathbf{G}^\text{int}$, respectively.  
The objective is to learn a projection function $f(\mathbf{G}^\text{sem}, \mathbf{G}^\text{int}) \rightarrow y$, where $y \in \{0, 1\}$ is the ground-truth label indicating whether the news is real or fake.

\noindent \textbf{Semantic graph notations.}
To extract semantic clues, unlike prior works \citep{karimi2019learning, xiao2024msynfd} that treat news narratives as a whole, we analyze news articles at the sentence level to capture local narrative structures.
In addition, we incorporate global relationships among sentences by using entities as references, enabling long-range contextual interactions and facilitating joint learning with news intent.
By integrating both local sentence-sentence and global sentence-entity relations, the semantic graph captures holistic semantic representations that reflect the narrative flow and preserve the logical and temporal progression of the article.
Specifically, the semantic heterogeneous graph $\mathbf{G}^\text{sem}$ is formulated as:
\begin{equation}
    \mathbf{G}^\text{sem} = \{ \mathbb{V}^\text{sem},\; \mathbb{L}^\text{sem} \}, \quad \mathbb{L}^\text{sem} = \mathbb{L}^{\text{local}} \cup \mathbb{L}^{\text{global}},
\end{equation}
where $\mathbb{V}^\text{sem} = \{ \{s_i\}_{i=1}^{m},\; \{e_j\}_{j=1}^{n} \}$ denotes the node set consisting of $m$ sentence and $n$ entity nodes.  
The edge set $\mathbb{L}^\text{sem}$ comprises local sentence-sentence $\mathbb{L}^{\text{local}}$ and global sentence-entity edges $\mathbb{L}^{\text{global}}$.

\noindent \textbf{Intent graph notations.}
To comprehensively capture intent representations, \modelname designs a coarse-to-fine strategy that combines high-level intent abstraction with fine-grained modeling.
Specifically, we first leverage LLMs to reason the coarse-grained news intent from $k$ conceptual perspectives
(\textit{e.g.,} belief, desire, etc. \citep{wang2025exploring}) to drive the coarse-grained intent nodes $\mathbb{C}$ within the node set $\mathbb{V}^\text{int}$ of the heterogeneous intent graph, serving as global anchors that represent the overarching intent of the news article.
To further capture how such intents are conveyed throughout the news narrative, we introduce $(k\times l)$ learnable fine-grained intent nodes $\mathbb{F}$ within $\mathbb{V}^\text{int}$, which are dynamically updated during training to model detailed intent implementation, based on the correlations between the coarse-grained intent signals and narrative characteristics.
This hierarchical design allows the intent graph to encode both high-level intent structures and their concrete realizations, enabling more expressive and adaptive intent modeling.
Concretely, the heterogeneous intent graph $\mathbf{G}^\text{int}$ is defined as:
\begin{equation}
    \mathbf{G}^\text{int} = \{ \mathbb{V}^\text{int},\; \mathbb{L}^\text{int} \}, \quad 
    \mathbb{L}^\text{int} = \mathbb{L}^{\text{coarse}} \cup \mathbb{L}^{\text{fine}},
\end{equation}
where $\mathbb{V}^\text{int} = \{ \mathbb{C},\; \mathbb{F} \}$ denotes the node set, comprising $k$ coarse-grained intent nodes and $l$ fine-grained nodes for each coarse-grained node, totally obtaining $(k+k\times l)$ nodes.
The edge set $\mathbb{L}^\text{int}$ includes coarse-coarse links $\mathbb{L}^{\text{coarse}}$ and coarse-fine links $\mathbb{L}^{\text{fine}}$.

\subsection{Semantic Graph Initialization} \label{sec:graph_constructon_sem}
To split news articles into sentences and extract corresponding entities, we utilize the SpaCy library for news content processing, then a pre-trained language model (PLM) is adopted to initialize the node embeddings $\mathbf{H}^\text{sem}$ for each node in $\mathbb{V}^\text{sem}$:

\begin{equation}
    \mathbf{H}^\text{sem} = \text{PLM}(\mathbb{V}^\text{sem}),
\end{equation}
where $\mathbf{H}^\text{sem} \in \mathbb{R}^{{(m+n)} \times d}$, consisting of $m$ sentence embeddings and $n$ entity embeddings with the embedding dimension of $d$.

For the initialization of edges within the semantic graph, we consider both local $\mathbb{L}^{\text{local}}$ and global $\mathbb{L}^{\text{global}}$ relationships between sentences and entities.
To model the narrative flow between sentences, previous works typically use full connections between sentence nodes \citep{hu2021compare}, resulting in overwhelming the model with redundant information or limiting the receptive field.
Instead, \modelname utilizes a sliding window-based connection strategy for sentence-sentence edges $\mathbb{L}^{\text{local}}$, which balances the computational efficiency and contextual awareness of local structures.
Specifically, sentences \( s_i \) and \( s_j \) are connected if their positions in the news fall within a predefined window size \( w \), as follows:
\begin{equation}
    \mathbb{L}^{\text{local}} = \{(s_i, s_j) \mid |i - j| \leq w\}.
\end{equation}
This ensures that each sentence is directly connected to a manageable number of neighboring sentences, providing sufficient local context to model its role in the narrative structure while maintaining computational efficiency.
Beyond local sentence relationships, we leverage entity information to incorporate long-distance context interaction, modeling global relationships guided by entity-sentence connections.
For sentence \( s_i \) and entity \( e_j \), an edge is added in $\mathbf{G}_\text{sem}$ if \( e_j \) is mentioned in \( s_i \) as:
\begin{equation}
    \mathbb{L}^{\text{global}} = \{(s_i, e_j) \mid e_j \in \text{Entities}(s_i)\}.
\end{equation}
These edges capture shared semantic contexts among sentences centered on the same entities, enabling the graph to effectively model interactions between narrative components.

\subsection{Intent Graph Initialization} \label{sec:graph_constructon_int}
\modelname proposes a coarse-to-fine strategy to initialize the intent graph $\mathbf{G}^\text{int}$.
The coarse-grained intent nodes $\mathbb{C}$ are initialized using a frozen generative language model with news content and a set of prompts $\mathbf{q} = \{q_1, \dots, q_k\}$ corresponding to the intent analysis framework with $k$ analytical perspectives (\textit{e.g.,} belief, desire, etc.).
Being framework-agnostic, this process enables the flexible incorporation of evolving theoretical intent analysis frameworks, guided by interdisciplinary progress.
Formally, given a news text $t$, the embedding of coarse-grained intent nodes is:
\begin{equation}
\begin{aligned}
    \mathbb{C} = \text{Generator}(t,\mathbf{q}), \quad
    \mathbf{H}^\text{c}=\text{PLM}(\mathbb{C}),
\end{aligned}
\end{equation}
where $\mathbf{H}^\text{c} \in \mathbb{R}^{k \times d}$ with feature dimension of $d$.

While these coarse-grained nodes provide high-level perspectives of news intent, they ignore the correlation between intent and narrative characteristics, limiting the ability to capture specific nuances in different news articles.
To address this limitation, we introduce several learnable fine-grained intent nodes.
Specifically, given the $i$-th coarse-grained intent node embedding $\mathbf{h}^{\rm c}_{i} \in \mathbf{H^{\rm c}}$, we initialize the corresponding fine-grained intent node embeddings $\{\mathbf{h}^{\rm f}_{ij}\}_{j=1}^{l} \in \mathbf{H^{\rm f}}$, as follows:

\begin{equation}
\begin{aligned}
    \mathbf{h}^{\rm f}_{ij} \leftarrow \mathbf{h}^{\rm f}_{ij} + \text{softmax}& \left({\left(\mathbf{h}^{\rm c}_{i} + \mathbf{h}^{\rm f}_{ij}\right) \cdot \mathbf{H}^\text{sem}_{\rm sen}}^T\right) \cdot \mathbf{H}^\text{sem}_{\rm sen},\\
\end{aligned}
\end{equation}
where $\mathbf{H}^\text{sem}_{\rm sen}$ represents the sentence part within the semantic node embeddings $\mathbf{H}^\text{sem}$ and $\text{softmax}(\cdot)$ represents softmax function.

For the edge initialization of $\mathbf{G}^\text{int}$, \modelname incorporates both coarse-to-coarse and coarse-to-fine relationships, denoted as $\mathbb{L}^{\text{coarse}}$ and $\mathbb{L}^{\text{coarse}}$, respectively.
$\mathbb{L}^{\text{fine}}$ encodes logical dependencies among high-level intent aspects defined by the intent analysis framework (\textit{e.g.,} \textit{belief} $\rightarrow$ \textit{desire}).
On the other hand, since each coarse-grained intent embedding contributes to the computation of its corresponding fine-grained intent nodes, $\mathbb{L}^{\text{fine}}$ establishes full connections between each coarse-grained node and its associated fine-grained intent nodes.
In this way, \modelname models both the high-level intent aspects and the correlation between intent and narrative characteristics, providing comprehensive representations of news intents.

\subsection{Dual-level Graph Update}
After the initialization of $\mathbf{G}^\text{sem}$ and $\mathbf{G}^\text{int}$, we perform dual-level graph updating to extract semantic and intent representations, based on the aforementioned graph nodes and edges.
The proposed dual-level graph updating considers two kinds of message passing among connected graph nodes, consisting of (i) local message passing, which captures the dependencies between nodes based on learnable edge weights and aggregates information from neighboring nodes, and (ii) global message passing that updates node embeddings using global context through a super root node, ensuring that a global perspective influences all nodes.

\noindent \textbf{Local message passing.} In this process, for each node $v \in \mathbb{V}^\text{sem} \cup \mathbb{V}^\text{int}$, $v$ exchange its own messages $\mathbf{h}_v \in \mathbb{H}^\text{sem} \cup \mathbb{H}^\text{c} \cup \mathbb{H}^\text{f}$ with its neighbors $\{\mathbf{h}_u|u \in \mathcal{N}(v)\}$ through learnable edge weights, where $\mathcal{N}(v)$ denotes the neighbors of node $v$.
To obtain the message passing intensity to each node, we adaptively compute the edge weights for each edge, based on the content of the source and target nodes and their difference. Specially, edge feature $\mathbf{e}_{uv}$ is computed as:
\begin{equation}
\mathbf{e}_{uv} = \mathbf{h}_u\, \| \, \mathbf{h}_v\, \| \, {\rm abs}(\mathbf{h}_u - \mathbf{h}_v), \quad u \in \mathcal{N}(v),
\end{equation}
where $\mathbf{h}_u$ and $\mathbf{h}_v$ are the node embeddings of the source and target nodes, ${\rm abs}(\cdot)$ is the absolute difference function, and $\, \| \,$ denotes vector concatenation.
Then, the edge weight $w_{uv}$ can be computed by passing $\mathbf{e}_{uv}$ through a Sigmoid function to normalize it:
\begin{equation}
w_{uv} = \sigma(\mathbf{W}_{\text{e}} \mathbf{e}_{uv} + b_{\text{e}}),
\end{equation}
where $\sigma$ denotes the Sigmoid function, and $\mathbf{W}_{\text{e}}$ and $b_{\text{e}}$ are the learnable metric and bias term. 
The resulting edge weight $w_{uv}$ determines the strength of the connection between nodes $u$ and $v$.
Finally, local message passing is performed to aggregate and update each node embedding of $\mathbf{G}^\text{sem}$ and $\mathbf{G}^\text{int}$, as follows:
\begin{equation}
    \mathbf{h}_v^{(l)} = \mathbf{W}_1 \mathbf{h}_v^{(l-1)} + \mathbf{W}_2 \sum_{u \in \mathcal{N}(v)} w_{uv} \cdot \mathbf{h}_u^{(l-1)},
\end{equation}
where $\mathbf{h}_v^{(l)}$ is the updated embedding of node $v$ at $l$-th layer, $\mathbf{h}_v^{(l-1)}$ is the current embedding at $(l-1)$-th layer and $\mathbf{W}_1, \mathbf{W}_2$ are learnable weight matrices. 

\noindent \textbf{Global message passing.}
Purely local message passing is limited by its inability to model global coherence, which is crucial for understanding the holistic graph structure.
Although stacking multi-layer local message passing can get a larger receptive field, it may result in over-smoothing \citep{oonograph} and over-squashing \citep{alonbottleneck}.

To complement the missing global information in the node embedding, we incorporate a global message passing strategy by introducing a super root node into the graph structure, which serves as a central aggregator that encapsulates the global information of the input graph.
Specifically, we represent the super root node as a learnable embedding \(\mathbf{h}^{\text{root}} \in \mathbb{R}^{1 \times d} \) that is randomly initialized and optimized during training. This dynamic representation allows the super root node to adapt to the specific information and structural properties of the graph.

For the node embeddings $\mathbf{h}_v$, its contribution to global narrative can be computed via a linear transformation followed by a softmax mechanism. At layer $l$, the super root node is updated as:
\begin{equation}
\mathbf{h}^{\text{root}} \leftarrow \mathbf{h}^{\text{root}} + \sum_v\text{softmax}(\mathbf{W}_\text{r}\mathbf{h}_{v}^{(l-1)} + b_\text{r}) \cdot \mathbf{h}_{v}^{(l-1)},
\end{equation}
where $\mathbf{W}_\text{r}$ and $b_\text{r}$ denote a learnable weight matrix and bias term, respectively.  
This step enables the root node to aggregate information from all other nodes in the graph in a weighted manner.

To integrate the updated global graph information $\mathbf{h}^{\text{root}}$ back into the graph, we apply a residual fusion with the previous layer’s node embeddings:
\begin{equation}
\mathbf{h}_{v}^{(l)} = \psi(\mathbf{h}^{\text{root}}) + \mathbf{h}_{v}^{(l-1)},
\end{equation}
where $\psi(\cdot)$ is a non-linear transformation function.
By leveraging the super root node and its dynamic interaction with graph nodes, through this global message-passing mechanism, nodes that represent sentences, entities, or intents are not only informed by their immediate neighbors but are also contextually enriched by the global embedding, allowing the semantic and intent representations to capture both detailed and overarching narratives.

\subsection{Dynamic Pathway-based Graph Alignment} \label{sec:space_alignment}
After obtaining the graph-based representations for semantics and intent, an intuitive approach for joint learning is to directly connect the nodes in the intent and semantic graphs and perform interaction between each node. However, this straightforward operation overlooks the representation gap between intents and semantics and could bring quadratically redundant computational costs regarding node numbers. 
To this end, we propose a dynamic pathway-based graph alignment module, leveraging pseudo nodes to perform bidirectional message passing and aggregation within a common space.

\noindent\textbf{Pseudo nodes and edges.}
To achieve this goal, we introduce pseudo nodes and edges to interact and align the representations between semantic and intent graphs.
Specifically, pseudo nodes act as conceptual bridges, facilitating effective interactions between semantic and intent signals by building bidirectional (\textit{i.e.}, semantic-pseudo-intent and intent-pseudo-semantic) message passing.

In practice, the pseudo node set $\mathbb{V}^\text{p}$ contains $r$ pseudo nodes, and the corresponding node embeddings $\mathbf{H}^\text{p} \in \mathbb{R}^{r \times d}$ are randomly initialized with the feature dimension $d$.
Together with the pseudo edges, which fully connect the semantic and intent graph nodes with the pseudo nodes, we transform the previous nodes within the semantic and intent graphs into a common space:
\begin{equation}
    \begin{aligned}
        \mathbb{V}^\text{com} & = \mathbb{V}^\text{sem} \cup \mathbb{V}^\text{int} \cup \mathbb{V}^\text{p}, \\
        \mathbb{L}^\text{com} &= \{(p, v) \mid p \in \mathbb{V}^\text{p},\;v \in \mathbb{V}^\text{sem} \cup \mathbb{V}^\text{int}\}.
    \end{aligned}
\end{equation}

\noindent\textbf{Dynamic message passing.}
Building upon the common space, we further design a dynamic message-passing mechanism to ensure that \modelname captures the nuanced relationships between heterogeneous graph nodes, enabling the bidirectional message passing and aggregation, while eliminating redundant node connections, thus resulting in aligned and effective deception representations.
Specifically, for a pseudo edge between $p$ and $v$, we first introduce its edge feature $\mathbf{e}_{pv}$ using a multi-layer
perceptron (MLP), incorporating node types ($v, p$) and edge direction ($p\rightarrow v$) context, to guide the message passing process, as follows:
\begin{equation}
    \mathbf{e}_{pv} = \text{MLP}\left(\text{oneHot}(p) \, \| \, \text{oneHot}(v) \, \| \, \text{oneHot}(p\rightarrow v) \right),
\end{equation}
where \(\text{oneHot}(\cdot)\) denotes the one-hot encoding of the source/target node type and the edge direction.
This enables \modelname to model nuanced relationships among nodes based on their semantic or intent roles, which is essential for capturing the subtle dependencies that may indicate misleading narratives.

Specifically, based on edge features, \modelname employs an attention mechanism to assign dynamic weights to pseudo edges, thereby identifying the most critical information pathways and facilitating reasoning over deceptive clues.  
For each pseudo node $p$, its embedding $\mathbf{h}_p \in \mathbf{H}^{\rm p}$ is dynamically updated based on the proximity to its neighboring nodes in the graph as follows:
\begin{equation}
\begin{aligned}
\mathbf{h}_p^{(l)} &= {\rm MLP}\left( \mathbf{h}_p^{(l-1)} + \sum_{(p, v) \in \mathbb{L}^\text{com}} \alpha_{pv} \cdot \mathbf{h}_v^{(l-1)}\right) , \\
\alpha_{pv} &= \text{softmax}\left((\mathbf{h}_p^{(l-1)} + \mathbf{e}_{pv}) \cdot (\mathbf{h}_v^{(l-1)} + \mathbf{e}_{pv})\right),
\end{aligned}
\end{equation}
where $\mathbf{h}_p^{(l)}$ denotes the updated embedding of node $p$ at the $l$-th layer, and $\mathbf{h}_p^{(l-1)}$ is the embedding from the previous layer. $\mathbf{e}_{pv}$ represents the feature of the edge $(p,v)$, and $\alpha_{pv}$ is the attention weight computed over the connected edge set $\mathbb{L}^\text{com}$.

During the alignment process, pseudo nodes ensure effective information exchange by acting as intermediaries between the semantic and intent graphs.
By dynamically adjusting edge attributes, critical information is strengthened while irrelevant or redundant information is gradually discarded, thus enhancing the model's capacity to extract complementary features from both semantic and intent signals and ultimately improving its ability to detect subtle indicators of fake news.

\subsection{Model Prediction and Optimization}
To perform fake news detection, we leverage the pseudo nodes as information bottlenecks that selectively filter out irrelevant content while emphasizing critical interactions.  
We aggregate features from the common space by applying mean pooling over the embeddings of the pseudo nodes:
\begin{equation}
\mathbf{h}_\text{cls} = \frac{1}{r} \sum_{\mathbf{h}_{p}\in \mathbf{H}^{\rm p}} \mathbf{h}_{p}.
\end{equation}

The pooled representation $\mathbf{h}_\text{cls}$ is then fed into an MLP to generate the prediction:
\begin{equation}
\hat{y} = \text{MLP}(\mathbf{h}_\text{cls}).
\end{equation}

Finally, the model is optimized using the cross-entropy loss:
\begin{equation}
\mathcal{L}_\text{cls} = - y_i \log(\hat{y}_i) - (1 - y_i) \log(1 - \hat{y}_i),
\end{equation}
where $y_i \in \{0, 1\}$ is the ground-truth label for the $i$-th instance.

\section{Experiments}

In this section, we present empirical results to demonstrate the effectiveness of our proposed method \modelname. These experiments are designed to answer the following experimental questions (EQs): 
\begin{itemize}[nosep,leftmargin=1em,labelwidth=*,align=left]
    \item[\textbf{EQ1} ]How does \modelname perform compared with state-of-the-art fake news detection methods?
    \item[\textbf{EQ2} ]How effective is the framework design of \modelname?
    \item[\textbf{EQ3} ]How do hyperparameters affect model performance?
\end{itemize}

\subsection{Experimental Setup}
\subsubsection{Datasets}
We conduct extensive experiments on four datasets from diverse languages and sources, including \textbf{PolitiFact}, \textbf{GossipCop}, \textbf{Weibo}, and \textbf{LLMFake}.
PolitiFact is an English dataset provided by \citet{popat2018declare}, which is collected based on the political fact-checking website. GossipCop is an English dataset based on the FakeNewsNet repository \citep{Shu_FakeNewsNet}, featuring news discussed on English social media. Weibo \citep{sheng2022zoom} is a Chinese dataset containing news from multiple domains spread on the Chinese social media platform Weibo. LLMFake \citep{Chen_LLM_Mis} is an LLM-generated fake news dataset proposed to combat fake news in the age of LLMs.
Following \citep{zhu2022generalizing}, we adopt a more practical and challenging evaluation protocol by splitting the train-validation-test set chronologically to simulate real-world scenarios where the writing patterns and themes of news evolve over time\footnote{Except for LLMFake, we performed a random split due to the absence of actual timestamps for the generated samples.}. Data statistics are shown in \tablename~\ref{tab:dataset}. 

\begin{table}[htbp]\small
\centering
\renewcommand{\arraystretch}{1.1}
    \setlength\abovecaptionskip{6pt}
    \setlength\belowcaptionskip{6pt}
\caption{\label{tab:dataset} Statistics of the datasets used in our experiments.}
  \setlength{\tabcolsep}{4pt}
  \begin{tabular}{lrrrrrrr}
  \toprule
  \multicolumn{1}{l}{\multirow{2}[4]{*}[0.3em]{\textbf{Dataset}}} & \multicolumn{2}{c}{\multirow{1}[1]{*}[0.2em]{\textbf{Train}}} & \multicolumn{2}{c}{\multirow{1}[2]{*}[0.3em]{\textbf{Validation}}} & \multicolumn{2}{c}{\textbf{Test}} & \multicolumn{1}{c}{\multirow{2}[4]{*}[0.1em]{\textbf{Total}}} \\
\cmidrule(lr){2-3}  \cmidrule(lr){4-5} \cmidrule(lr){6-7}
& \multicolumn{1}{l}{Fake} & \multicolumn{1}{l}{Real} & \multicolumn{1}{l}{Fake} & \multicolumn{1}{l}{Real} & \multicolumn{1}{l}{Fake} & \multicolumn{1}{l}{Real} \\
  \midrule
  \textbf{PolitiFact} 
        & 1,117 & 1,295 & 169 & 175 & 367 & 323 & 3,446\\
  \textbf{GossipCop} 
        & 2,001 & 4,951 & 593 & 1,731 & 594 & 1,710 & 11,580\\
  \textbf{Weibo} 
        & 2,497 & 7,455 & 497 & 1,910 & 745 & 2,937 & 16,041\\
  \textbf{LLMFake} 
        & 408   & 289   & 118 & 80    & 59  & 41    & 995\\
  \bottomrule
  \end{tabular}%
  \setlength\textfloatsep{-16pt}
\end{table}%

\subsubsection{Baselines} 
We include two groups of methods for comparison. The first group is intent-agnostic detection methods:
\begin{itemize}[leftmargin=*,itemsep=2pt,topsep=0pt,parsep=0pt]
    \item \textit{BERT} \citep{devlin_BERT}, a pre-trained language model that is widely used as the text encoder for fake news detection~\cite{fakebert, zhu2022generalizing, Nan_MDFEND}, with the last layer finetuned conventionally. 
    \item \textit{EANN} \citep{Wang_EANN}, which aims to learn event-invariant representations for fake news detection to mitigate potential news event biases arising in temporal scenarios.
    \item \textit{MDFEND} \citep{Nan_MDFEND}, integrating the mixture of experts to capture the domain information of news to address the decline in detection performance caused by potential domain transfer.
    \item \textit{BERT-Emo} \citep{zhang2021mining}, considering the emotions conveyed by publishers of the news piece for fake news detection, including emotional signals such as emotion category, emotional lexicon, emotional intensity, and sentiment score.
    \item \textit{ENDEF} \citep{zhu2022generalizing}, introduces an entity debiasing framework to mitigate the temporal bias introduced by entities within news.
    \item \textit{MGIN-AG}~\cite{MGINAG}, a structure-aware model which processes news text using a dependency tree and jointly employs BERT and GCN to generate augmented features for fake news detection.
    \item \textit{MSynFD} \citep{xiao2024msynfd}, a structure-aware model that builds a multi-hop syntactic dependency graph to model syntax information and sequentially aware semantic information for fake news detection.

    \item  \textit{LLM}, to validate the performance of LLM in the fake news detection task, we prompt an LLM to make veracity judgments based on the provided news content.
    \item  \textit{GenFEND} \citep{nan2024let}, a generated feedback-enhanced detection framework, which generates news comments by prompting LLMs with diverse user profiles and aggregating them from multiple subpopulation groups for fake news detection. 
\end{itemize}
The second group is intent-aware fake news detection methods: 
\begin{itemize}[leftmargin=*,itemsep=2pt,topsep=0pt,parsep=0pt]
    \item \textit{DMInt} \citep{wang2025exploring}, which trains a news intent feature extractor using human-labeled intent data to derive intent features. These features are then combined with semantic features through MLPs.
    \item \textit{DM-Inter} \citep{wang2024misinformation} adopts a self-supervised training strategy, where a generative model is guided to produce intent-related binary classification responses and is optimized using its own sharpened predictions. The generator’s parameters serve as intent features, then fused with news embeddings for veracity judgment.
    \item \textit{LLM-Int}, we incorporate perspective-specific prompts to guide the LLM in analyzing the news intent. The resulting analyses are then prompted to produce the final veracity judgment.
\end{itemize}

EANN and MDFEND are implemented following \citet{zhu2022generalizing}.
For other baseline methods, we adopt the official settings used in the corresponding paper.
For \textit{LLM} and \textit{LLM-Int}, two well-recognized commercial LLMs are exploited to make a competitive comparison. Specifically, we prompt \textit{GPT-4o-mini} \citep{chatgpt} for English data, while for Chinese data, we utilize \textit{GLM-4-plus} \citep{GLM4}, a competitive counterpart for the Chinese language following prior work \citep{nan2024let}.
The prompt used is listed in Prompt 1 (the \dashuline{underlined text} is only for \textit{LLM-Int}). 
\begin{tcolorbox}[title=Prompt 1:  Prompt for Veracity Judgment, boxrule=0pt, left=1mm, right=1mm, top=1mm, bottom=1mm, fontupper=\small] \label{prompt}
    \textbf{System Prompt:} Given the following news piece \dashuline{and the corresponding intent analyses}, predict the veracity of this news piece. If the news piece is more likely to be fake, return 1; otherwise, return 0. Please refrain from providing ambiguous assessments such as undetermined. \\
    \textbf{Context Prompt:} News: [\textit{$t$}] \dashuline{; analyses: [\textit{news intent analyses}]}. The answer (Arabic numerals) is:
\end{tcolorbox}

\begin{sidewaystable} \scriptsize
    \centering
    \renewcommand{\arraystretch}{1.25}
    \setlength\tabcolsep{1pt}
    \setlength\dbltextfloatsep{-6pt}
    \caption{Experimental results of our proposed method \modelname across four datasets. The best and the second-best results are \textbf{bolded} and \underline{underlined}, respectively. The {$\pm$} values denote the standard deviation, * indicates 0.05 significance level from a paired t-test comparing \modelname with the best baseline method.}
    \begin{tabular}{lcccccccccc}
    \toprule
        \multicolumn{1}{c}{\multirow{2}[4]{*}[0.3em]{\textbf{Method}}} 
        & \multicolumn{5}{c}{\textbf{Dataset: PolitiFact}} & \multicolumn{5}{c}{\textbf{Dataset: GossipCop}} \\
        \cmidrule(lr){2-6} \cmidrule(lr){7-11}  
        & macF1 & Acc & AUC & F1$_{\text{real}}$ & F1$_{\text{fake}}$ 
        & macF1 & Acc & AUC & F1$_{\text{real}}$ & F1$_{\text{fake}}$ \\
        \midrule

        BERT
            & 0.6041{\scriptsize \color{grayv} $\pm$.0087} 
            & 0.6083{\scriptsize \color{grayv} $\pm$.0083} 
            & 0.6476{\scriptsize \color{grayv} $\pm$.0035} 
            & 0.6199{\scriptsize \color{grayv} $\pm$.0032} 
            & 0.6087{\scriptsize \color{grayv} $\pm$.0156}
            
            & 0.7875{\scriptsize \color{grayv} $\pm$.0070}
            & 0.8439{\scriptsize \color{grayv} $\pm$.0043}
            & 0.8780{\scriptsize \color{grayv} $\pm$.0066}
            & 0.8973{\scriptsize \color{grayv} $\pm$.0091}
            & 0.6781{\scriptsize \color{grayv} $\pm$.0100}
        \\
        EANN
            & 0.6107{\scriptsize \color{grayv} $\pm$.0161} 
            & 0.6146{\scriptsize \color{grayv} $\pm$.0145} 
            & 0.6463{\scriptsize \color{grayv} $\pm$.0015} 
            & 0.6159{\scriptsize \color{grayv} $\pm$.0080} 
            & 0.6207{\scriptsize \color{grayv} $\pm$.0252}

            & 0.7931{\color{grayv} \scriptsize $\pm$.0054}
            & 0.8518{\color{grayv} \scriptsize $\pm$.0071}
            & 0.8778{\color{grayv} \scriptsize $\pm$.0114}
            & 0.9040{\color{grayv} \scriptsize $\pm$.0058}
            & 0.6823{\color{grayv} \scriptsize $\pm$.0108}
        \\
        MDFEND
            & 0.6195{\scriptsize \color{grayv} $\pm$.0027} 
            & 0.6218{\scriptsize \color{grayv} $\pm$.0031} 
            & 0.6510{\scriptsize \color{grayv} $\pm$.0082} 
            & 0.5968{\scriptsize \color{grayv} $\pm$.0048} 
            & 0.6461{\scriptsize \color{grayv} $\pm$.0008}

            & 0.7905{\scriptsize \color{grayv} $\pm$.0084}
            & 0.8518{\scriptsize \color{grayv} $\pm$.0071}
            & 0.8712{\scriptsize \color{grayv} $\pm$.0063}
            & 0.9037{\scriptsize \color{grayv} $\pm$.0038}  
            & 0.6772{\scriptsize \color{grayv} $\pm$.0088}
        \\
        BERT-Emo
            & 0.6251{\scriptsize \color{grayv} $\pm$.0034} 
            & 0.6272{\scriptsize \color{grayv} $\pm$.0041} 
            & 0.6607{\scriptsize \color{grayv} $\pm$.0116} 
            & 0.6148{\scriptsize \color{grayv} $\pm$.0178} 
            & 0.6353{\scriptsize \color{grayv} $\pm$.0154}
            
            & 0.7912{\scriptsize \color{grayv} $\pm$.0030}  
            & 0.8455{\scriptsize \color{grayv} $\pm$.0051}
            & 0.8800{\scriptsize \color{grayv} $\pm$.0100}
            & 0.8974{\scriptsize \color{grayv} $\pm$.0060}
            & 0.6849{\scriptsize \color{grayv} $\pm$.0044}
        \\
        ENDEF
            & 0.6189{\scriptsize \color{grayv} $\pm$.0018} 
            & 0.6284{\scriptsize \color{grayv} $\pm$.0128} 
            & 0.6635{\scriptsize \color{grayv} $\pm$.0056} 
            & 0.6357{\scriptsize \color{grayv} $\pm$.0096} 
            & 0.6129{\scriptsize \color{grayv} $\pm$.0082}
            
            & 0.8010{\scriptsize \color{grayv} $\pm$.0088}
            & 0.8520{\scriptsize \color{grayv} $\pm$.0101}
            & 0.8855{\scriptsize \color{grayv} $\pm$.0080}
            & 0.9020{\scriptsize \color{grayv} $\pm$.0124}
            & 0.6987{\scriptsize \color{grayv} $\pm$.0094}
        \\
        MGIN-AG
            & 0.6228{\scriptsize \color{grayv} $\pm$.0035} 
            & 0.6251{\scriptsize \color{grayv} $\pm$.0037} 
            & 0.6544{\scriptsize \color{grayv} $\pm$.0117} 
            & 0.6002{\scriptsize \color{grayv} $\pm$.0072} 
            & 0.6494{\scriptsize \color{grayv} $\pm$.0047}
            
            & 0.8072{\scriptsize \color{grayv} $\pm$.0051}
            & 0.8593{\scriptsize \color{grayv} $\pm$.0047}
            & 0.8916{\scriptsize \color{grayv} $\pm$.0050}
            & 0.9074{\scriptsize \color{grayv} $\pm$.0072}
            & 0.7069{\scriptsize \color{grayv} $\pm$.0066}
        \\
        MSynFD
            & 0.6261{\scriptsize \color{grayv} $\pm$.0090} 
            & 0.6252{\scriptsize \color{grayv} $\pm$.0046} 
            & 0.6543{\scriptsize \color{grayv} $\pm$.0108} 
            & 0.6002{\scriptsize \color{grayv} $\pm$.0018} 
            & {0.6528}{\scriptsize \color{grayv} $\pm$.0092}            
            
            & 0.8094{\scriptsize \color{grayv} $\pm$.0071}
            & \uline{0.8702}{\scriptsize \color{grayv} $\pm$.0032}
            & 0.8961{\scriptsize \color{grayv} $\pm$.0051}
            & \uline{0.9153}{\scriptsize \color{grayv} $\pm$.0054}
            & 0.7103{\scriptsize \color{grayv} $\pm$.0088}
        \\
        LLM
            & 0.4133{\scriptsize \color{grayv} $\pm$.0008} 
            & 0.5471{\scriptsize \color{grayv} $\pm$.0014} 
            & 0.5188{\scriptsize \color{grayv} $\pm$.0013} 
            & 0.1331{\scriptsize \color{grayv} $\pm$.0003} 
            & \uline{0.6935}{\scriptsize \color{grayv} $\pm$.0013}
            
            & 0.6641{\scriptsize \color{grayv} $\pm$.0031}
            & 0.7135{\scriptsize \color{grayv} $\pm$.0022}
            & 0.6895{\scriptsize \color{grayv} $\pm$.0020}
            & 0.7930{\scriptsize \color{grayv} $\pm$.0030}
            & 0.5352{\scriptsize \color{grayv} $\pm$.0025}
        \\
        GenFEND
            & \uline{0.6380}{\scriptsize \color{grayv} $\pm$.0034} 
            & {0.6383}{\scriptsize \color{grayv} $\pm$.0021} 
            & \uline{0.6736}{\scriptsize \color{grayv} $\pm$.0059} 
            & \uline{0.6405}{\scriptsize \color{grayv} $\pm$.0062} 
            & {0.6366}{\scriptsize \color{grayv} $\pm$.0041}
            
            & \uline{0.8367}{\scriptsize \color{grayv} $\pm$.0064}
            & {0.8598}{\scriptsize \color{grayv} $\pm$.0033}
            & \uline{0.9110}{\scriptsize \color{grayv} $\pm$.0078}
            & {0.8995}{\scriptsize \color{grayv} $\pm$.0101}
            & \uline{0.7302}{\scriptsize \color{grayv} $\pm$.0080}
        \\
        \cmidrule{1-11}
        DMInt
            & 0.6346{\scriptsize \color{grayv} $\pm$.0073}
            & \uline{0.6406}{\scriptsize \color{grayv} $\pm$.0053}
            & 0.6670{\scriptsize \color{grayv} $\pm$.0088}
            & 0.5880{\scriptsize \color{grayv} $\pm$.0090}
            & {0.6812}{\scriptsize \color{grayv} $\pm$.0061}
            
            & 0.8001{\scriptsize \color{grayv} $\pm$.0034}
            & 0.8589{\scriptsize \color{grayv} $\pm$.0040}
            & 0.8984{\scriptsize \color{grayv} $\pm$.0052}
            & 0.9086{\scriptsize \color{grayv} $\pm$.0024}
            & 0.6914{\scriptsize \color{grayv} $\pm$.0037}
        \\
        DM-inter
            & 0.6177{\scriptsize \color{grayv} $\pm$.0087}
            & 0.6151{\scriptsize \color{grayv} $\pm$.0100}
            & 0.6574{\scriptsize \color{grayv} $\pm$.0108}
            & 0.6320{\scriptsize \color{grayv} $\pm$.0065}
            & 0.5966{\scriptsize \color{grayv} $\pm$.0088}
            
            & 0.7960{\scriptsize \color{grayv} $\pm$.0091}
            & 0.8524{\scriptsize \color{grayv} $\pm$.0110}
            & 0.8670{\scriptsize \color{grayv} $\pm$.0087}
            & 0.9033{\scriptsize \color{grayv} $\pm$.0054}
            & 0.6886{\scriptsize \color{grayv} $\pm$.0106}
        \\
        LLM-Int
            & 0.4392{\scriptsize \color{grayv} $\pm$.0044} 
            & 0.5696{\scriptsize \color{grayv} $\pm$.0041} 
            & 0.5555{\scriptsize \color{grayv} $\pm$.0041} 
            & 0.3211{\scriptsize \color{grayv} $\pm$.0055} 
            & 0.6574{\scriptsize \color{grayv} $\pm$.0033}

            & 0.6833{\scriptsize \color{grayv} $\pm$.0037} 
            & 0.6849{\scriptsize \color{grayv} $\pm$.0042} 
            & 0.7308{\scriptsize \color{grayv} $\pm$.0057} 
            & 0.6944{\scriptsize \color{grayv} $\pm$.0161} 
            & 0.6775{\scriptsize \color{grayv} $\pm$.0086}
        \\
        \rowcolor{lightgrayv}\textbf{\modelname}
        & \textbf{0.6823}{\scriptsize \color{grayv} $\pm$.0053}* 
        & \textbf{0.6855}{\scriptsize \color{grayv} $\pm$.0036}* 
        & \textbf{0.7321}{\scriptsize \color{grayv} $\pm$.0095}* 
        & \textbf{0.6505}{\scriptsize \color{grayv} $\pm$.0060}* 
        & \textbf{0.7141}{\scriptsize \color{grayv} $\pm$.0079}*

        & \textbf{0.8653}{\scriptsize \color{grayv} $\pm$.0024}*
        & \textbf{0.9002}{\scriptsize \color{grayv} $\pm$.0021}*
        & \textbf{0.9285}{\scriptsize \color{grayv} $\pm$.0042}*
        & \textbf{0.9338}{\scriptsize \color{grayv} $\pm$.0039}*
        & \textbf{0.7968}{\scriptsize \color{grayv} $\pm$.0020}*
        \\
    \hline
    \specialrule{0em}{0.5pt}{0.5pt}
    \hline
    \specialrule{0em}{0.pt}{3pt}
        \multicolumn{1}{c}{\multirow{2}[4]{*}[0.3em]{\textbf{Method}}} 
        & \multicolumn{5}{c}{\textbf{Dataset: Weibo}} & \multicolumn{5}{c}{\textbf{Dataset: LLMFake}} \\
        \cmidrule(lr){2-6} \cmidrule(lr){7-11}  
        & macF1  & Acc  & AUC  & F1$_{\text{real}}$ & F1$_{\text{fake}}$ 
        & macF1  & Acc  & AUC  & F1$_{\text{real}}$ & F1$_{\text{fake}}$ \\
        \midrule

        BERT
            & 0.7550{\scriptsize \color{grayv} $\pm$.0034} 
            & 0.8354{\scriptsize \color{grayv} $\pm$.0031} 
            & 0.8580{\scriptsize \color{grayv} $\pm$.0027} 
            & 0.8904{\scriptsize \color{grayv} $\pm$.0013} 
            & 0.6124{\scriptsize \color{grayv} $\pm$.0018}

            & 0.8670{\scriptsize \color{grayv} $\pm$.0081} 
            & 0.8690{\scriptsize \color{grayv} $\pm$.0109} 
            & 0.9434{\scriptsize \color{grayv} $\pm$.0033} 
            & 0.8428{\scriptsize \color{grayv} $\pm$.0122} 
            & 0.8906{\scriptsize \color{grayv} $\pm$.0043}
        \\
        EANN
            & 0.7271{\color{grayv} \scriptsize $\pm$.0036} 
            & 0.8297{\color{grayv} \scriptsize $\pm$.0053}  
            & 0.8051{\color{grayv} \scriptsize $\pm$.0095}
            & 0.8775{\color{grayv} \scriptsize $\pm$.0076}
            & 0.5998{\color{grayv} \scriptsize $\pm$.0045}
            
            &   -     &    -    &    -    &    -   &  -
        \\
        MDFEND
            & 0.7194{\scriptsize \color{grayv} $\pm$.0123}
            & 0.7789{\scriptsize \color{grayv} $\pm$.0061}
            & 0.8303{\scriptsize \color{grayv} $\pm$.0106}
            & 0.8518{\scriptsize \color{grayv} $\pm$.0055}
            & 0.5592{\scriptsize \color{grayv} $\pm$.0063}
            
            &   -     &    -    &    -    &    -   &  -
        \\
        BERT-Emo
            & 0.7584{\scriptsize \color{grayv} $\pm$.0101}
            & 0.8438{\scriptsize \color{grayv} $\pm$.0079}
            & 0.8742{\scriptsize \color{grayv} $\pm$.0073}
            & 0.9017{\scriptsize \color{grayv} $\pm$.0068}
            & 0.6155{\scriptsize \color{grayv} $\pm$.0090}

            & 0.8797{\scriptsize \color{grayv} $\pm$.0127} 
            & 0.8731{\scriptsize \color{grayv} $\pm$.0106} 
            & 0.9401{\scriptsize \color{grayv} $\pm$.0105} 
            & 0.8708{\scriptsize \color{grayv} $\pm$.0083} 
            & 0.9087{\scriptsize \color{grayv} $\pm$.0071}
            
        \\
        ENDEF
            & 0.7683{\scriptsize \color{grayv} $\pm$.0049}
            & 0.8579{\scriptsize \color{grayv} $\pm$.0052}
            & 0.8836{\scriptsize \color{grayv} $\pm$.0121}
            & 0.9024{\scriptsize \color{grayv} $\pm$.0069}
            & 0.6242{\scriptsize \color{grayv} $\pm$.0035}

            & 0.8825{\scriptsize \color{grayv} $\pm$.0039} 
            & 0.8871{\scriptsize \color{grayv} $\pm$.0022} 
            & 0.9439{\scriptsize \color{grayv} $\pm$.0033} 
            & 0.8687{\scriptsize \color{grayv} $\pm$.0034} 
            & 0.9055{\scriptsize \color{grayv} $\pm$.0043}
                   
        \\       
        MGIN-AG
            & 0.7744{\scriptsize \color{grayv} $\pm$.0037}
            & 0.8659{\scriptsize \color{grayv} $\pm$.0033}
            & 0.8741{\scriptsize \color{grayv} $\pm$.0047}
            & 0.9183{\scriptsize \color{grayv} $\pm$.0082}
            & 0.6314{\scriptsize \color{grayv} $\pm$.0100}

            & 0.8848{\scriptsize \color{grayv} $\pm$.0025} 
            & 0.8881{\scriptsize \color{grayv} $\pm$.0036} 
            & 0.9352{\scriptsize \color{grayv} $\pm$.0042} 
            & 0.8491{\scriptsize \color{grayv} $\pm$.0037} 
            & 0.9058{\scriptsize \color{grayv} $\pm$.0036}
        \\
        MSynFD
            & 0.7793{\scriptsize \color{grayv} $\pm$.0099}
            & \uline{0.8685}{\scriptsize \color{grayv} $\pm$.0071}
            & 0.8806{\scriptsize \color{grayv} $\pm$.0112}
            & \uline{0.9268}{\scriptsize \color{grayv} $\pm$.0110}
            & 0.6382{\scriptsize \color{grayv} $\pm$.0060}
            
            & 0.8932{\scriptsize \color{grayv} $\pm$.0072} 
            & {0.8957}{\scriptsize \color{grayv} $\pm$.0111} 
            & {0.9441}{\scriptsize \color{grayv} $\pm$.0059} 
            & 0.8768{\scriptsize \color{grayv} $\pm$.0104} 
            & 0.9058{\scriptsize \color{grayv} $\pm$.0085}
            
        \\
        LLM
            & 0.6824{\scriptsize \color{grayv} $\pm$.0033} 
            & 0.7317{\scriptsize \color{grayv} $\pm$.0028} 
            & 0.7700{\scriptsize \color{grayv} $\pm$.0069} 
            & 0.8075{\scriptsize \color{grayv} $\pm$.0044} 
            & 0.5573{\scriptsize \color{grayv} $\pm$.0031}
            
            & 0.6589{\scriptsize \color{grayv} $\pm$.0022} 
            & 0.6667{\scriptsize \color{grayv} $\pm$.0018} 
            & 0.6612{\scriptsize \color{grayv} $\pm$.0047} 
            & 0.6076{\scriptsize \color{grayv} $\pm$.0035} 
            & 0.7103{\scriptsize \color{grayv} $\pm$.0038}
        \\
        GenFEND
            & \uline{0.7869}{\scriptsize \color{grayv} $\pm$.0039}
            & 0.8595{\scriptsize \color{grayv} $\pm$.0064}
            & \uline{0.8853}{\scriptsize \color{grayv} $\pm$.0077}
            & 0.9189{\scriptsize \color{grayv} $\pm$.0056}
            & 0.6510{\scriptsize \color{grayv} $\pm$.0112}
            
            & \uline{0.9006}{\scriptsize \color{grayv} $\pm$.0041} 
            & 0.8956{\scriptsize \color{grayv} $\pm$.0091} 
            & \uline{0.9458}{\scriptsize \color{grayv} $\pm$.0067} 
            & {0.8705}{\scriptsize \color{grayv} $\pm$.0027} 
            & \uline{0.9109}{\scriptsize \color{grayv} $\pm$.0016}

        \\
        \cmidrule{1-11}
        DMInt
            &   -     &    -    &    -    &    -   &  -
            & 0.8994{\scriptsize \color{grayv} $\pm$.0074}
            & \uline{0.9032}{\scriptsize \color{grayv} $\pm$.0081}
            & 0.9388{\scriptsize \color{grayv} $\pm$.0089}
            & \uline{0.8800}{\scriptsize \color{grayv} $\pm$.0071}
            & {0.9089}{\scriptsize \color{grayv} $\pm$.0035}
        \\
        DM-inter
            & 0.7763{\scriptsize \color{grayv} $\pm$.0054}
            & 0.8623{\scriptsize \color{grayv} $\pm$.0077}
            & 0.8777{\scriptsize \color{grayv} $\pm$.0112}
            & 0.9150{\scriptsize \color{grayv} $\pm$.0098}
            & 0.6236{\scriptsize \color{grayv} $\pm$.0114}
            
            & 0.8534{\scriptsize \color{grayv} $\pm$.0113}
            & 0.8602{\scriptsize \color{grayv} $\pm$.0088}
            & 0.8698{\scriptsize \color{grayv} $\pm$.0121}
            & 0.8219{\scriptsize \color{grayv} $\pm$.0079}
            & 0.8850{\scriptsize \color{grayv} $\pm$.0110}
        \\
        LLM-Int
            & 0.5392{\scriptsize \color{grayv} $\pm$.0062} 
            & 0.5696{\scriptsize \color{grayv} $\pm$.0058} 
            & 0.5555{\scriptsize \color{grayv} $\pm$.0058} 
            & 0.4211{\scriptsize \color{grayv} $\pm$.0078} 
            & \uline{0.6574}{\scriptsize \color{grayv} $\pm$.0046}
            
            & 0.6958{\scriptsize \color{grayv} $\pm$.0040}
            & 0.6886{\scriptsize \color{grayv} $\pm$.0086}
            & 0.7299{\scriptsize \color{grayv} $\pm$.0026}
            & 0.7121{\scriptsize \color{grayv} $\pm$.0048}
            & 0.6806{\scriptsize \color{grayv} $\pm$.0033}
        \\
        \rowcolor{lightgrayv}\textbf{\modelname}
        & \textbf{0.8055}{\scriptsize \color{grayv} $\pm$.0031}*
        & \textbf{0.8807}{\scriptsize \color{grayv} $\pm$.0029}*
        & \textbf{0.9102}{\scriptsize \color{grayv} $\pm$.0048}*
        & \textbf{0.9319}{\scriptsize \color{grayv} $\pm$.0065}*
        & \textbf{0.6841}{\scriptsize \color{grayv} $\pm$.0079}*

        & \textbf{0.9218}{\scriptsize \color{grayv} $\pm$.0042}* 
        & \textbf{0.9247}{\scriptsize \color{grayv} $\pm$.0050}* 
        & \textbf{0.9567}{\scriptsize \color{grayv} $\pm$.0044}* 
        & \textbf{0.9067}{\scriptsize \color{grayv} $\pm$.0020}* 
        & \textbf{0.9369}{\scriptsize \color{grayv} $\pm$.0065}*
        \\
    \bottomrule
    \end{tabular}
    \label{tab:main_experiment}
\end{sidewaystable}

\subsubsection{Implementation Details} \label{sssec:implementation_details}
We utilize AdamW as the optimizer, the batch size is 64, and the learning rate is $2e-4$. The maximum number of entities is 32, the embedding dimension is 256, the graph neural network depth is set to 3, and the MLP hidden sizes are $[128, 128]$.
We adopt \textit{bert-base-uncased} and \textit{bert-base-chinese} as text encoders for English and Chinese data, respectively, with the last layer conventionally fine-tuned, consistent with the baseline settings. Additionally, we utilize \textit{Qwen2.5-7B-Instruct} and \textit{Llama-3.1-8B-Instruct} as generators to initialize coarse-grained intent nodes in Section~\ref{sec:graph_constructon_int} for Chinese and English data, respectively, with the model parameters frozen. The different choices of semantic encoders and coarse-level intent generators will be analyzed in Section~\ref{sec:extend_sem_int_encoder}.
Moreover, we select $4$ conceptual perspectives, including belief, plan, desire and outcome, for coarse-grained news intent initialization, \textit{i.e.}, $k = 4$ in Section~\ref{sec:graph_constructon_int}, following the conceptual intent framework proposed in \citep{wang2025exploring}.

\subsubsection{Evaluation Metrics} To evaluate the models' performance, following most existing works \citep{Wang_EANN, shu2019defend, zhu2022generalizing, xiao2024msynfd}, Area Under ROC (AUC), accuracy (Acc), macro F1 score (macF1) and the F1 scores of fake and real class (F1$_{\text{fake}}$ and F1$_{\text{real}}$) are reported.

\subsection{Overall Performance Comparison (EQ1)}
\label{sec:exp4.2}
The average results of \modelname and the compared methods across four datasets over three runs are shown in \tablename~\ref{tab:main_experiment}. Since the LLMFake dataset lacks domain labels, \textit{EANN} and \textit{MDFEND} methods cannot be applied to this dataset. Additionally, the feature extractor of \textit{MEInt} was trained on English corpora, making it unsuitable for processing Weibo dataset. We have the following key observations: 

\begin{itemize}[leftmargin=*,itemsep=2pt,topsep=0pt,parsep=0pt]  
    \item \textbf{Consistent performance gain.} \modelname consistently outperforms all competitive methods across all datasets, including those leveraging more powerful LLMs.
    The improvements are measured against the best-performing baseline on each individual metric and dataset, underscoring the robustness and generalizability of \modelname across diverse evaluation scenarios.
    
    \item \textbf{Exceeding recent state-of-the-art intent-enhanced models.} \modelname not only outperforms semantic-based baselines but also surpasses recently proposed intent-enhanced methods.
    For example, compared to the state-of-the-art intent-enhanced methods DMInt \citep{wang2025exploring} and DM-Inter \citep{wang2024misinformation} that adopt fixed classification-based intent modeling without alignment, our \modelname consistently achieves much better performance on all datasets, \textit{e.g.}, $0.8653$ \textit{v.s.} $0.7960$ of DM-inter and $0.8001$ of DMInt for macF1 on the GossipCop dataset.
    This is mainly because the unstructured coarse-to-fine intent modeling strategy and effective alignment between intent and semantic signals enable the capture of more diverse intent manifestations across varied news contexts, while mitigating their inherent representational differences, ultimately leading to superior results.
    
    \item \textbf{\modelname excels at identifying the fake news class.} Our method achieves outstanding results on the F1$_{\text{fake}}$ metric, with relative improvements of up to 9.12\% on the GossipCop datasets. This highlights the practical value of \modelname, as accurate recognition of the fake class is especially critical for reducing the spread of misinformation in real-world applications.

\end{itemize}

\subsection{Effectiveness of \modelname's Design (EQ2)}

As semantic and intent graph learning, along with the graph alignment module, are central to our proposed method, we conduct the following experiments to investigate their effectiveness.

\begin{table}[htbp]\small
  \centering
  \renewcommand{\arraystretch}{1.2}
  \caption{Performance comparison of \modelname and its variants across four datasets. The best results are \textbf{bolded}. \label{tab:ablation}}
    \setlength\tabcolsep{1.1pt}
    \begin{tabular}{l cc cc cc cc}
    \toprule
    \multicolumn{1}{c}{\multirow{2}[4]{*}[0.3em]{\textbf{Method}}} 
    & \multicolumn{2}{c}{\textbf{PoltiFact}} 
    & \multicolumn{2}{c}{\textbf{GossipCop}} 
    & \multicolumn{2}{c}{\textbf{Weibo}} 
    & \multicolumn{2}{c}{\textbf{LLMFake}} \\
    \cmidrule(lr){2-3} \cmidrule(lr){4-5} \cmidrule(lr){6-7}  \cmidrule(lr){8-9}
    & \multicolumn{1}{c}{macF1} & \multicolumn{1}{c}{Acc} 
    & \multicolumn{1}{c}{macF1} & \multicolumn{1}{c}{Acc}
    & \multicolumn{1}{c}{macF1} & \multicolumn{1}{c}{Acc}
    & \multicolumn{1}{c}{macF1} & \multicolumn{1}{c}{Acc}\\
    \midrule   
    \rowcolor{lightgrayv}\modelname 
        & 0.6823 & 0.6855
        & \textbf{0.8653} & \textbf{0.9002} 
        & \textbf{0.8055} & \textbf{0.8807}  
        & \textbf{0.9218} & \textbf{0.9247}
        \\
    \; \textit{w/o} Entity 
        & 0.6609  & 0.6667
        & 0.8549  & 0.8932
        & 0.7848  & 0.8666 
        & 0.8925  & 0.8954
        \\
    \; \textit{w/o} Window
        & 0.6619  & 0.6623
        & 0.8578  & 0.8954 
        & 0.7968  & 0.8520 
        & 0.9015  & 0.9063
        \\
    \; \textit{w/o} $\text{Int}_\text{f}$
        & 0.6626  & 0.6637
        & 0.8432  & 0.8759
        & 0.7957  & 0.8675
        & 0.9039 &  0.9062
        \\
    \; \textit{w/o} Global
        & 0.6594  & 0.6595
        & 0.8537  & 0.8889
        & 0.7991  & 0.8762 
        & 0.9015  & 0.9032
        \\
    \; \textit{w/o} DPGA
        & 0.6449  & 0.6448
        & 0.8374  & 0.8772 
        & 0.7902  & 0.8580 
        & 0.8935  & 0.8958
        \\
    \cmidrule{1-9} 
    \modelname ($\mathbb{C}'$)
        & \textbf{0.6828}  & \textbf{0.6913}
        & 0.8611  & 0.8958  
        & 0.8018  & 0.8768   
        & 0.9143  & 0.9167
        \\    
    \bottomrule
    \end{tabular}%
  \setlength\dbltextfloatsep{-6pt}
\end{table}%

\subsubsection{Effectiveness of Semantic and Intent Graph Learning}

To investigate the design of intent and semantic graph learning, we conduct the following studies:
\begin{itemize}[leftmargin=*,itemsep=2pt,topsep=0pt,parsep=0pt]
    \item \modelname (\textit{w/o} Entity), which removes the entity nodes, retaining only the sentence nodes in the semantic graph.
    \item \modelname (\textit{w/o} Window), which eliminates the edge connections based on the window design and employs a fully connected approach for sentence connections within the semantic graph.
    \item \modelname (\textit{w/o} $\text{Int}_\text{f}$), which removes the fine-grained intent nodes, leaving only coarse intent nodes in the intent graph.
    \item \modelname (\textit{w/o} Global), which removes the global message passing module, relying solely on local message passing.
\end{itemize}

As shown in \tablename~\ref{tab:ablation}, we find that: 
(i) Entity nodes facilitate long-distance context interaction by modeling the global narrative structure through entity-sentence connections. Removing them (\textit{w/o} Entity) impairs this capacity and leads to performance drops.
(ii) Sliding window-based sentence interaction effectively captures moderate contextual semantics, providing stable local context modeling.
(iii) Fine-grained intent learning improves the model's capability by strengthening the correlation between intent and specific narrative characteristics of news. Its absence (\textit{w/o} $\text{Int}_\text{f}$) results in performance degradation;
(iv) Global message-passing mechanism enables the effective capture of global news context; its removal (\textit{w/o} Global) weakens the model's capability.

\subsubsection{Effectiveness of Dynamic Pathway-based Graph Alignment Module}
To investigate the advantages of the proposed graph alignment module with pseudo nodes and edges, we replace it with an intuitive approach (denoted as \textit{w/o} {DPGA} in Table~\ref{tab:ablation}) by directly connecting the nodes in the intent and semantic graphs, followed by a fully connected interaction between them.
From the result, one can see that the performance significantly drops without the proposed graph alignment module, indicating that \modelname effectively captures bidirectional dependencies between intent and semantic signals and emphasizes critical interactions while discarding irrelevant ones, which is essential for fake news detection.

\subsubsection{Stability of \modelname With Different Coarse-grained News Intent Analysis Framework \label{ssec:ab_intent_frame}}
To evaluate the stability of \modelname to different intent analysis framework, we substitute the original theoretical framework with four intent perspectives $\mathbb{C} \in \mathbb{V}^\text{int}$ with an alternative one \citep{wang2024misinformation}, which conceptualizes intent from nine perspectives (\textit{e.g.,} public, individual, connection, etc.), denoted as \modelname($\mathbb{C}'$).
As shown in \tablename~\ref{tab:ablation}, \modelname($\mathbb{C}'$) demonstrates stable performance across all datasets, consistently outperforming all baseline methods while showing slight variations compared to the default framework. This stability suggests that our fine-grained intent modeling strategy effectively bridges high-level intent concepts to specific news implementations, maintaining robustness across different theoretical foundations. These results indicate that \modelname's framework design is adaptable to various intent conceptualizations, while also highlighting the potential for exploring more optimal intent analysis frameworks in future work.

\subsection{Hyperparameter Analysis (EQ3)} \label{sec:hyper_parameter}

\begin{figure}
\setlength\abovecaptionskip{6pt}
\setlength\belowcaptionskip{6pt}
    \centering 
    \includegraphics[width=.8\linewidth]{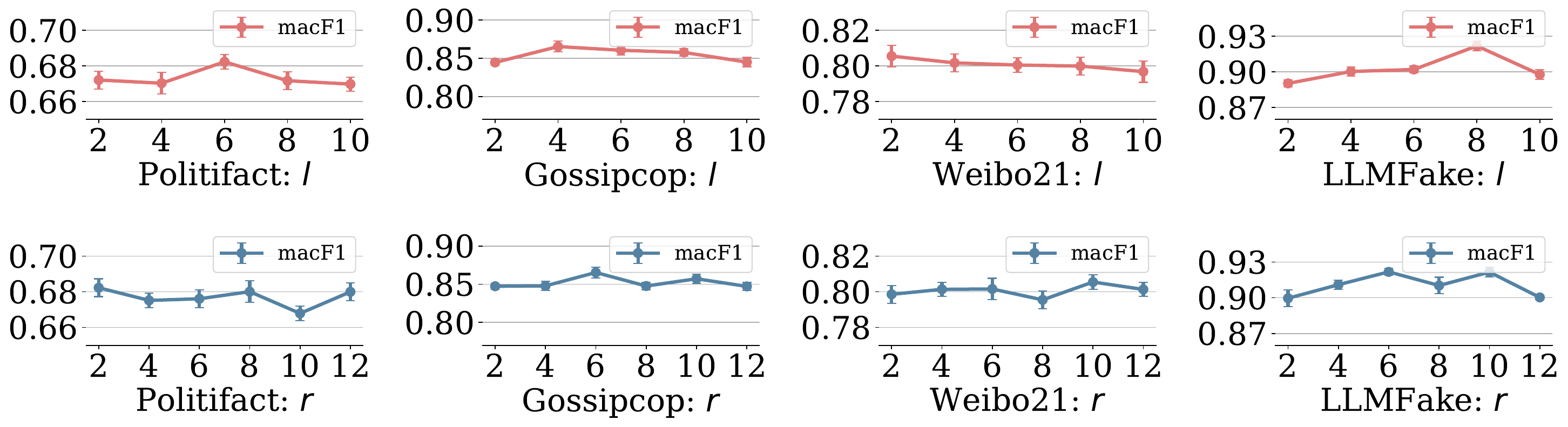}
    \caption{Impact of hyperparameters $l$ and $r$ on macF1.}\label{fig:hyperparameters}
\end{figure}

The influence of two key hyperparameters, the number of fine-grained intent nodes ($l$) and the number of pseudo nodes ($r$), is shown in Figure~\ref{fig:hyperparameters}. For $l$, the results show a clear trend where increasing $l$ initially improves performance as more fine-grained intent nodes contribute to capturing nuanced intent patterns, but excessive refinement leads to overfitting or redundancy. 
For $r$, the results suggest relative robustness, reflecting the adaptability of the designed mechanism in providing dynamic pathways for graph alignment, empowering flexible message passing, and allowing effective operation across diverse graph connectivity patterns.

\section{Further Analysis}
\begin{figure}
    \centering 
    \includegraphics[width=.65\linewidth]{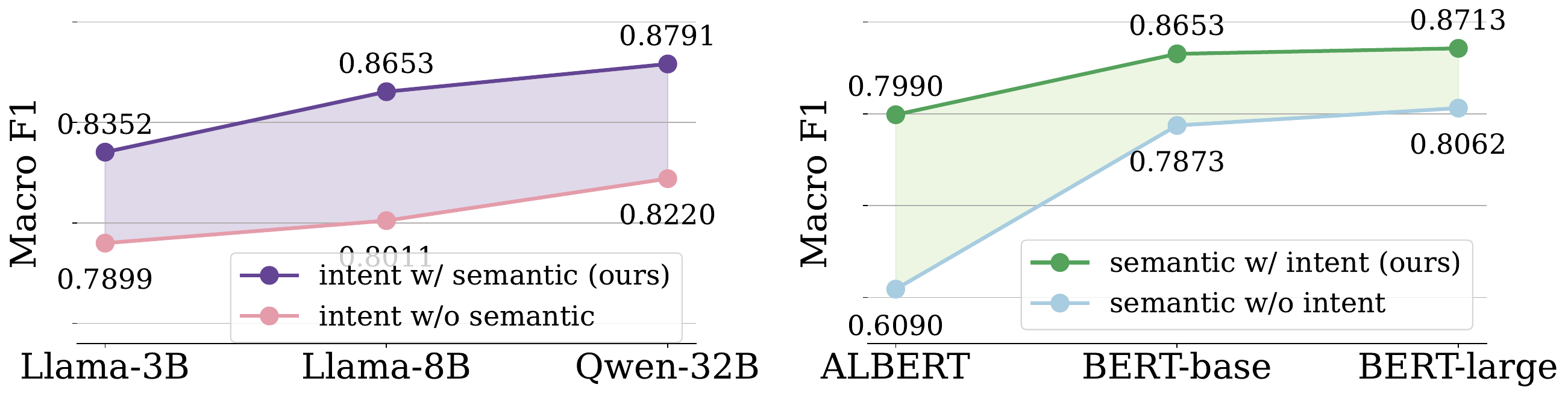}
    \caption{\label{fig:alignment} Model performance with and without intent-semantic joint learning across diverse semantic encoders and intent generators, showcasing framework extensibility.
    }
\end{figure}

\subsection{Extension Ability of \modelname} \label{sec:extend_sem_int_encoder}
To demonstrate the extensibility of the proposed method, we evaluate it under various model configurations on the GossipCop dataset by varying both semantic encoders and intent generators. As shown in Figure~\ref{fig:alignment}, our framework consistently yields performance gains across a range of model sizes—from lightweight models (\textit{e.g.,} ALBERT \citep{ALBERT} and LLaMA-3B \citep{dubey2024llama}), to standard settings used in our main experiments (\textit{e.g.,} BERT \citep{devlin_BERT} and LLaMA-8B), and further to larger-scale models (\textit{e.g.,} BERT-large and Qwen-32B \citep{qwen2})\footnote{Larger models are not included in this study due to resource constraints. Specifically, Llama-3.2-3B-Instruct, Llama-3.1-8B-Instruct, and Qwen2.5-32B-Instruct are used.}. These results demonstrate the superiority of the proposed joint learning framework in effectively leveraging the complementary strengths of semantics and intent.

These also illustrate the scalability of our method from cost-effective setups to high-performance scenarios: (i) it remains effective when paired with lighter modules, allowing for further reductions in computational cost; and (ii) it can integrate with stronger backbones to further boost performance.

\subsection{Case Analysis}
\begin{figure}[t]
\centering
    \includegraphics[width=.6\linewidth]{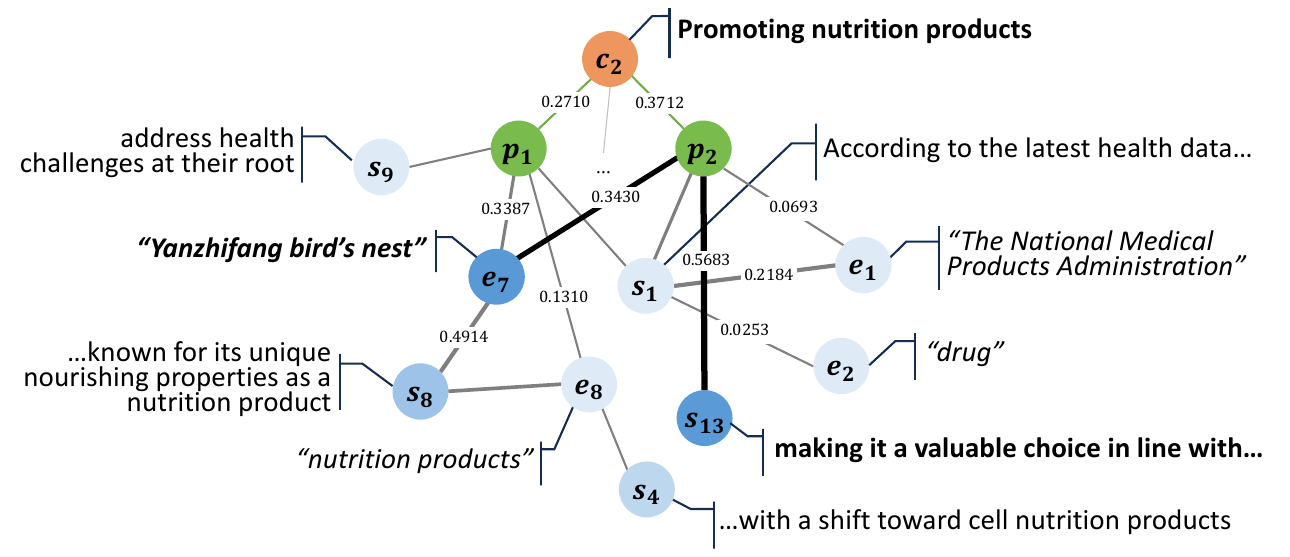}
    \caption{Case visualization analysis of a news item from the Weibo dataset after \textit{Graph Alignment}. Darker edges indicate critical information pathways. The pseudo nodes (in {\color{bettergreen}green}) act as dynamic bridges between semantic nodes (in {\color{bluev}blue}) and intent nodes (in {\color{myorange}orange}), enabling flexible message passing.}
    \label{fig:case}
\end{figure}

To illustrate \modelname’s capability in jointly learning semantics and intents for fake news detection, we analyze the example in \figurename~\ref{fig:case}. This example reveals how our method uncovers underlying intent and its connection to a deceptive narrative. The graph consists of {\color{black}semantic nodes ($s$, $e$)}, {\color{black}intent nodes ($c$)}, and {\color{black}pseudo nodes ($p$)}, linked by learned edges. For clarity, some nodes and edges are omitted.

Notably, entity nodes facilitate long-distance context interaction; for example, sentence nodes $s_4$ and $s_8$ connect via entity node $e_8$, enabling a shorter, 2-hop information propagation path that enhances the model’s capture of extended narrative structure.
Furthermore, semantic and intent nodes are able to interact through intermediate pseudo nodes.
During the graph learning process, critical information is strengthened while irrelevant or redundant information is gradually weakened. For example, $e_7$ and $s_{13}$ connect effectively through $p_2$, highlighting the intent of promoting nutrition products and exaggerating their effectiveness—key deception clues extracted by the model.
Without these dynamic pseudo node bridges, the distance between deceptive clues would increase, hindering the model’s ability to capture dynamic patterns indicative of fake news.

\section{Conclusion}
We presented \modelname, a fake news detection method that bridges intent and semantics to enhance deception understanding by modeling the crucial interplay between inherent intents and flexible semantics.
We pointed out that the joint learning of these signals is essential for enhancing fake news detection.
Furthermore, we proposed that the challenge of this joint learning approach lies in effectively leveraging their complementary advantages while bridging the representation gaps between them.
We first reformulated semantic and intent signals into heterogeneous graphs for better expressive power, then built a common space for cross-signal alignment, and finally drove dynamic pathway-based bidirectional message passing and aggregation, thereby obtaining aligned and effective deception representations for fake news detection.
Extensive experiments demonstrate the effectiveness of \modelname.
Moreover, this research highlights the dual role of news intents: they not only motivate news writing but, when effectively utilized, can also play a crucial role in detecting fake news.

\noindent\textbf{Limitations.} Despite these contributions, our work has limitations that present opportunities for future research: (i) Our LLM-based intent modeling could benefit from more advanced reasoning approaches \citep{guo2025deepseek, yang2025qwen3} to better capture complex intent manifestations in deceptive content.
(ii) Our method is not intended for detecting unintentional misinformation, \textit{e.g.,} cases without deception such as inadvertent errors, so the performance is unclear for such cases. (iii) While harmful intent can be realized through truthful content, we target fake news as the more harmful case; nevertheless, our framework could naturally extend to detect manipulative framing of truthful content, offering broader applicability in combating information manipulation.

\normalem

\end{document}